\documentclass[11pt,a4paper]{article}

\usepackage[margin=1in]{geometry}
\usepackage[utf8]{inputenc}
\usepackage[T1]{fontenc}
\usepackage{abstract}
\usepackage{amsmath,amssymb,amsfonts}
\usepackage{booktabs}
\usepackage{longtable}
\usepackage{graphicx}
\usepackage{adjustbox}
\usepackage{caption}
\usepackage{float}
\usepackage{rotating}
\usepackage{pdflscape}
\usepackage{appendix}
\usepackage[numbers,sort&compress]{natbib}
\bibpunct[, ]{[}{]}{,}{n}{,}{,}

\usepackage{siunitx}
\usepackage{array}
\usepackage{multirow}
\usepackage{hyperref}
\usepackage{bookmark}
\usepackage{authblk}
\usepackage{microtype}

\usepackage{setspace}
\title{\textbf{Kalimati Vegetable Price Index Forecasting with a Momentum Corrected Online Stacking Ensemble}}

\author[1]{Sahaj Raj Malla}
\affil[1]{Department of Mathematics, Kathmandu University, Dhulikhel, Nepal \\
          \vspace{0.5em}
          \texttt{sm03200822@student.ku.edu.np}}

\date{}

\begin{document}


\maketitle

\begin{abstract}
Forecasting agricultural commodity prices in emerging economies is difficult due to high volatility, frequent supply disruptions, and strong cultural influences on demand. This study introduces the Kalimati Vegetable Price Index (KVPI), a new inverse-volatility weighted composite index that aggregates 135 daily wholesale commodities from Kathmandu over ten years (2013--2023). By creating a stable macro-level signal, the KVPI reduces the noise inherent in modelling individual crops. A rich set of 64 causally valid features was developed, including festival lead-lag effects, rolling statistics, and calendar variables. Fourteen forecasting models spanning statistical, tree-based, deep learning, hybrid, and transformer architectures were rigorously evaluated across short (7-day), medium (14- and 30-day), and long-term (90-day) horizons. Tree-based ensembles proved notably robust, while classical statistical models and complex transformers struggled with the noisy dataset. The proposed Momentum-Corrected Online Stacking Ensemble achieved the strongest performance, yielding a Root Mean Square Error (RMSE) of 1.771, an exceptionally low Mean Absolute Percentage Error (MAPE) of 0.68\%, and explaining 84.5\% of the variance ($R^2 = 0.845$) at the 90-day horizon. This open-source pipeline provides policymakers and supply chain actors in Nepal and similar markets with a practical, reliable tool for anticipating price movements and strengthening food security.
\end{abstract}

\vspace{1em}

\newpage

\section{Introduction}
\label{sec:intro}

\subsection{Background and Motivation}
Agriculture forms the backbone of Nepal's economy, contributing approximately 24\%--25\% to national GDP and employing over 60\% of the country's workforce \cite{worldbank2023nepal, nepaleconomicSurvey208081}. The Kalimati Fruits and Vegetables Market in Kathmandu functions as the nation’s largest wholesale hub, serving as the central node for the distribution of both domestically produced and imported agricultural commodities. 

However, agricultural commodity prices in emerging markets such as Nepal exhibit extreme volatility. This instability arises from a combination of structural supply chain weaknesses, heavy dependence on monsoon rainfall, frequent climate-induced disruptions, and sharp, culturally driven demand surges associated with major festivals \cite{bellemare2015rising}. Such price fluctuations create substantial challenges for policymakers, wholesalers, retailers, and smallholder farmers, often resulting in inefficient resource allocation, heightened food inflation, increased post-harvest losses, and compromised food security.

Accurate price forecasting is therefore essential for enhancing market transparency, supporting evidence-based policy decisions, and improving supply chain resilience. Nevertheless, modelling agricultural prices in developing economies remains particularly difficult due to high-frequency noise in daily transaction data, frequent missing values, and the complex non-linear interactions that characterise these markets \cite{makridakis2018statistical}.

\subsection{Problem Statement}
A substantial portion of existing literature on agricultural price forecasting focuses on individual commodities in isolation. While informative, this atomistic approach is highly vulnerable to commodity-specific shocks, data sparsity, and idiosyncratic noise, ultimately limiting model generalisability and practical utility. Datasets from emerging markets further compound these issues through inconsistent reporting, overlapping entries, and irregular temporal coverage.

From a methodological perspective, the field has evolved from classical statistical models such as ARIMA toward increasingly sophisticated machine learning and deep learning architectures \cite{lim2021time}. Traditional statistical methods effectively capture linear dependencies but struggle to incorporate high-dimensional exogenous factors such as rolling market statistics and cultural calendar effects \cite{zhang2003time}. Conversely, state-of-the-art deep learning models, including transformers (e.g., PatchTST) \cite{nie2023time} and neural basis expansion models (e.g., NBEATSx) \cite{olivares2022neural}, have demonstrated strong performance on large, clean datasets. However, they frequently suffer from overfitting and poor generalisation when applied to moderately sized, noisy economic time series typical of developing economies.

Consequently, no single modelling paradigm—statistical, machine learning, or deep learning—has yet proven sufficient for robust, multi-horizon forecasting of agricultural prices in volatile emerging market contexts.

\subsection{Research Contributions}
This study addresses these limitations by developing a comprehensive, reproducible framework for agricultural price forecasting in Nepal. The primary contributions are as follows:

\begin{enumerate}
    \item \textbf{Development of the Kalimati Vegetable Price Index (KVPI):} We construct a novel, inverse-volatility weighted composite index aggregating 135 commodities over ten years (2013--2023). By smoothing commodity-specific noise, the KVPI provides a stable macro-level benchmark of the Nepalese wholesale vegetable market.
    \item \textbf{High-Dimensional Feature Engineering:} We engineer a rich set of 64 exogenous predictors while strictly preventing look-ahead bias. This feature set uniquely integrates autoregressive lags, rolling statistics, calendar effects, and carefully designed lead/lag windows around major Nepali festivals (Dashain, Tihar, Chhath, Holi, Teej, and Nepali New Year) to explicitly capture culturally induced market shocks.
    \item \textbf{Extensive Multi-Horizon Benchmarking:} We rigorously evaluate 14 forecasting architectures—including naive baselines, statistical models, tree-based ensembles, deep recurrent networks, hybrid approaches, and state-of-the-art transformers—across short (7-day), medium (14- and 30-day), and long-term (90-day) horizons.
    \item \textbf{Momentum-Corrected Online Stacking Ensemble:} We propose a Momentum-Corrected Online Stacking Ensemble that dynamically adjusts predictions using a rolling residual derivative (slope of recent forecast errors) to apply momentum corrections during periods of systematic bias. This meta-learner, which optimally blends tree-based models (e.g., ExtraTrees) with recurrent networks (e.g., GRU), delivers superior stability and accuracy, particularly over extended forecasting horizons, achieving an RMSE of 1.771, a MAPE of 0.684\%, and an $R^2$ of 0.845 at the 90-day horizon.
\end{enumerate}

\section{Related Work}
\label{sec:literature}

\subsection{Agricultural Commodity Price Forecasting}
Early research on agricultural commodity price forecasting relied predominantly on linear econometric models grounded in the Box-Jenkins methodology \cite{box2015time}. Autoregressive Integrated Moving Average (ARIMA) and its seasonal extension (SARIMA) became standard tools for capturing periodic harvest cycles and short-term autocorrelations \cite{gaddi2025application}. These models offered interpretability and performed adequately under stable conditions.

However, their core assumptions of stationarity, linearity, and constant variance are frequently violated in emerging markets. Wholesale prices in regions like South Asia are subject to abrupt structural breaks caused by monsoon variability, transport disruptions, policy interventions, and sharp demand surges during cultural festivals. While extensions such as Vector Autoregression (VAR) and ARIMAX permit the inclusion of exogenous variables, they remain limited in modelling complex non-linear interactions and volatility clustering inherent to these markets.

Recent studies have highlighted the limitations of traditional approaches in volatile agricultural settings, particularly in developing economies where data quality and external shocks pose additional challenges \cite{manogna2025enhancing, tran2023predicting}.

\subsection{Machine Learning and Deep Learning in Time Series Forecasting}
The shortcomings of classical statistical models have prompted a shift toward flexible, non-parametric machine learning techniques. Tree-based ensembles, such as Random Forest and Extreme Gradient Boosting (XGBoost) \cite{chen2016xgboost}, have demonstrated strong performance by transforming time-series problems into supervised learning tasks through lag embedding. These methods naturally accommodate high-dimensional heterogeneous features—including rolling statistics, calendar effects, and event-based indicators—without requiring data stationarity.

Deep learning architectures further advanced the field by directly modelling sequential dependencies. Recurrent neural networks, particularly Long Short-Term Memory (LSTM) \cite{hochreiter1997long} and Gated Recurrent Units (GRU), effectively mitigate vanishing gradient problems and capture longer-term temporal patterns. More recently, Transformer-based models \cite{wen2022transformers} and neural basis expansion architectures such as N-BEATS \cite{oreshkin2019n} and NBEATSx \cite{olivares2022neural} have shown promise in long-horizon forecasting. Variants like PatchTST \cite{nie2023time} excel on large-scale multivariate datasets by processing subsequence patches efficiently.

Despite these advances, empirical evidence indicates that highly parameterised deep learning models often struggle with moderately sized, noisy datasets typical of emerging-market agricultural time series. Such models are prone to overfitting and exhibit poor generalisation compared to simpler tree-based methods when training data is limited or exhibits high volatility \cite{chen2023long, theofilou2025predicting}.

\subsection{Ensemble and Hybrid Architectures}
Recognising the complementary strengths and weaknesses of individual paradigms, researchers have increasingly explored hybrid and ensemble approaches. A seminal hybrid framework proposed by Zhang \cite{zhang2003time} decomposes a time series into linear and non-linear components, applying ARIMA to the former and a neural network to the residuals. While conceptually elegant, such sequential hybrids risk error propagation when the initial linear model is misspecified \cite{khashei2010artificial}.

Stacking ensembles provides a more robust alternative. Introduced by Wolpert \cite{wolpert1992stacked}, stacking trains a meta-learner on the predictions of diverse base models, allowing the integration of complementary inductive biases. In time-series applications, however, static ensembles can be vulnerable to concept drift. Dynamic Ensemble Selection (DES) methods address this limitation by evaluating base-model competence over rolling validation windows and adaptively selecting or weighting models \cite{cruz2018dynamic}. This adaptability is particularly valuable for volatile agricultural markets, where the relative performance of individual forecasters can shift rapidly with changing economic and seasonal conditions.

Recent work has further emphasised the value of ensemble strategies in agricultural price forecasting, showing consistent gains in accuracy and stability over single-model approaches \cite{kaabi2025utilizing, wang2026promise}. Building upon these foundations, the present study introduces a Momentum-Corrected Online Stacking Ensemble tailored to emerging-market agricultural price indices. This approach incorporates a rolling residual derivative mechanism to dynamically correct for systematic bias during periods of high volatility and cultural demand shocks, addressing a key limitation observed in both static ensembles and standard hybrid architectures.

\section{Methodology}
\label{sec:methodology}

\subsection{Data Description}
\subsubsection{Data Source}
This study utilises daily wholesale price records from the Kalimati Fruits and Vegetables Market, the primary wholesale hub for agricultural produce in Kathmandu, Nepal. The dataset was obtained from Open Data Nepal \cite{kalimati_tarkari_dataset} and covers a continuous ten-year period from June 2013 to September 2023. The raw dataset comprises 280,939 transaction records across 136 unique commodities, spanning vegetables, fruits, spices, and fishery products. Each record includes six key fields: Commodity, Date, Unit, Minimum Price, Maximum Price, and Average Price (in Nepalese Rupees). The data is publicly available at \url{https://opendatanepal.com/datasets/kalimati-tarkari-dataset}.

\subsubsection{Data Cleaning}
An automated Python script performed price field cleaning (removing currency prefixes and delimiters), date format normalisation, and unit standardisation. Commodities were classified into a hierarchical taxonomy using keyword matching (e.g., Root and Tuber Vegetables $\rightarrow$ Root Vegetables). Overlapping records were resolved by defining a composite unique key \texttt{[Commodity, Date, Minimum, Maximum, Average]}, with exact duplicates removed to produce a clean, unified baseline dataset.

\subsection{Data Preprocessing}
The preprocessing pipeline was designed to be fully deterministic and idempotent, with every step programmatically logged to ensure complete reproducibility and auditability.

\subsubsection{Schema Normalisation and Duplicate Removal}
Price fields were converted to numeric values, with missing entries preserved as \texttt{NaN}. Commodity names and units were standardised to title case. For days with multiple entries for the same commodity, the most recent record was retained.

\subsubsection{Outlier Detection and Treatment}
Given the susceptibility of agricultural prices to short-term shocks, outliers were detected using a per-commodity Interquartile Range (IQR) approach combined with strict domain constraints. Critically, to prevent data leakage, the IQR bounds ($Q_1$ and $Q_3$) were computed exclusively on the training set (dates $\le$ June 30, 2022). These frozen bounds were subsequently applied to cap outliers across the entire dataset:
\begin{equation}
\text{lower} = \max\left(Q_1 - k \cdot \text{IQR},\ P_{\min}\right), \quad \text{upper} = \min\left(Q_3 + k \cdot \text{IQR},\ P_{\max}\right)
\end{equation}
where $k=2.5$, $P_{\min}=1.0$, and $P_{\max}=1000.0$ Rs/kg. Extreme values were winsorised (capped at the computed bounds) rather than removed, preserving temporal continuity. In total, 4,903 values were adjusted across the dataset.

\subsubsection{Missing Date Imputation}
To support continuous time-series analysis, each commodity was reindexed to a complete daily calendar from its first to last recorded date. Missing values were imputed using a two-stage strategy:
\begin{itemize}
    \item Forward-fill for short gaps (maximum \(7\) days).
    \item Bidirectional linear interpolation for longer absences.
\end{itemize}
To ensure zero target leakage, this interpolation was performed independently within the training and testing partitions. No interpolation was permitted to cross the train/test split boundary. This process generated \(121{,}599\) imputed daily records.

\subsubsection{Kalimati Vegetable Price Index (KVPI) Construction}
Forecasting individual commodities in isolation is prone to excessive noise and limited generalisability. To address this, we constructed the Kalimati Vegetable Price Index (KVPI) as a stable macro-level benchmark.

Commodities with fewer than 365 days of data were excluded, leaving 135 series. For each commodity $(c)$, a base mean price $\bar{P}_c^{\text{base}}$ was calculated from the first 30 trading days. Normalised prices were computed as:
\begin{equation}
P_{c,t}^{\text{norm}} = \frac{P_{c,t}}{\bar{P}_c^{\text{base}}} \times 100
\end{equation}
An inverse-volatility weighting scheme was applied to reduce the influence of highly erratic commodities:
\begin{equation}
w_c = \frac{1 / \text{CV}_c}{\sum_{j} 1 / \text{CV}_j}, \quad \text{CV}_c = \frac{\sigma_c^{\text{base}}}{\bar{P}_c^{\text{base}}}
\end{equation}
The daily KVPI is the weighted average of available normalised prices:
\begin{equation}
\text{KVPI}_t = \frac{\sum_{c \in C_t} w_c \cdot P_{c,t}^{\text{norm}}}{\sum_{c \in C_t} w_c}
\end{equation}
The resulting index spans 3,757 continuous days with a range of [90.1, 219.9], effectively capturing broad market trends while mitigating individual commodity shocks.

\begin{figure}[!htbp]
\centering
\includegraphics[width=0.85\textwidth]{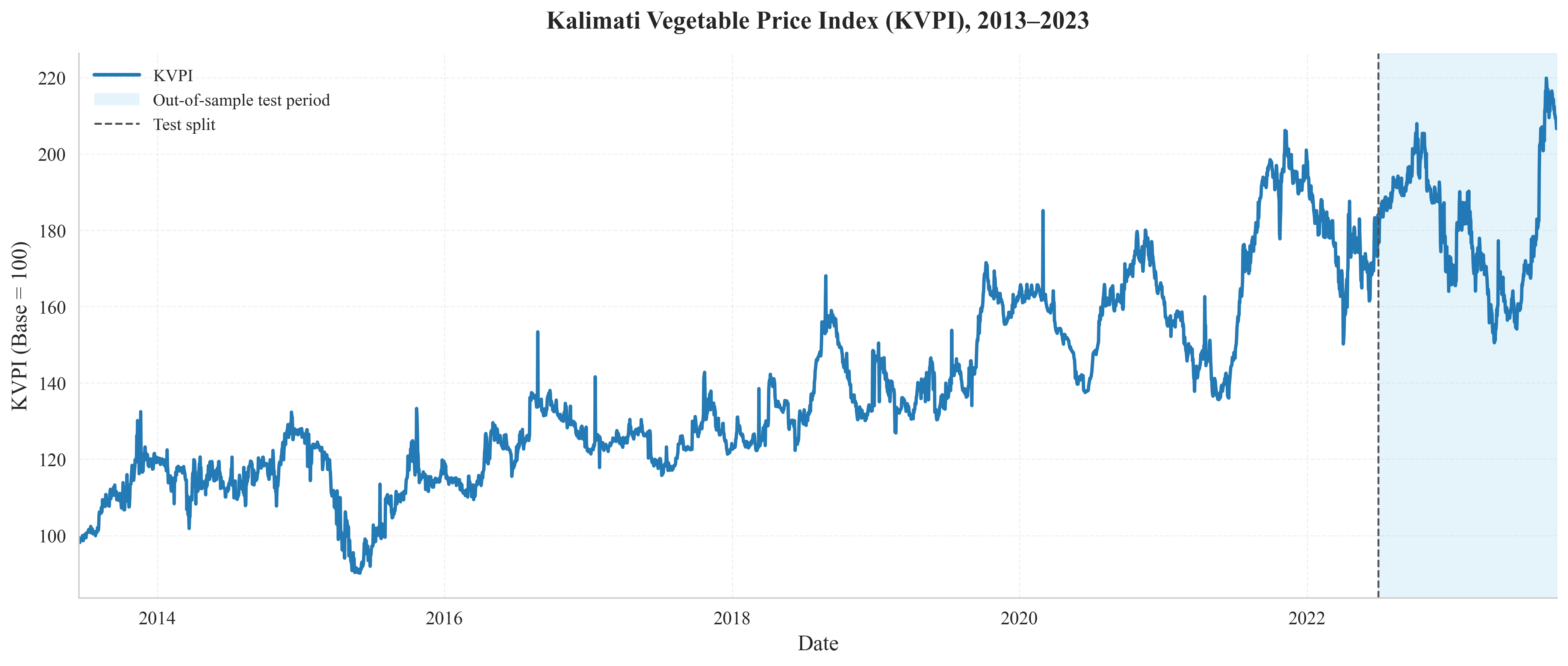}
\caption{Historical evolution of the Kalimati Vegetable Price Index (KVPI), 2013--2023, with the out-of-sample test period shaded.}
\label{fig:kvpi_price_series}
\end{figure}

\subsubsection{Series Diagnostics}
\begin{figure}[!htbp]
\centering
\includegraphics[width=0.85\textwidth]{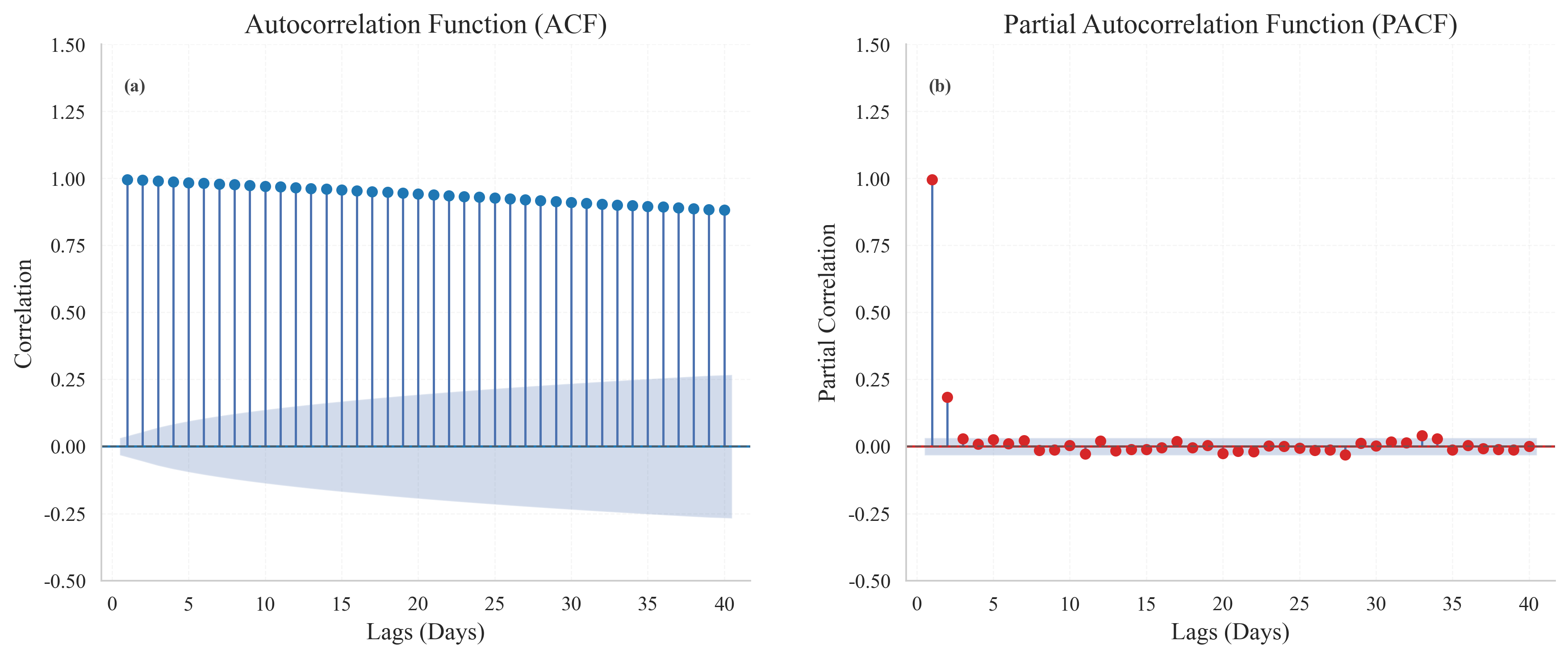}
\caption{Autocorrelation (ACF) and partial autocorrelation (PACF) plots of the KVPI.}
\label{fig:kvpi_acf_pacf}
\end{figure}

Figure~\ref{fig:kvpi_acf_pacf} shows the autocorrelation and partial autocorrelation structure of the KVPI. The ACF remains highly persistent over short and medium lags, while the PACF shows strong first-order dependence followed by rapid attenuation, motivating the use of lagged features and ARIMA-family baselines. To formally assess stationarity and justify the integration order for statistical modeling, an Augmented Dickey-Fuller (ADF) test was conducted. The test confirmed the presence of a unit root in the level series (test statistic = $-1.172$, $p = 0.686$), mathematically justifying the application of first-order differencing ($d=1$) in the subsequent ARIMA frameworks.

\subsection{Feature Engineering}
A high-dimensional feature matrix of 64 predictors was engineered to capture temporal, structural, and cultural dynamics. All lag- and rolling-based features were shifted backward by one day to strictly eliminate look-ahead bias. Furthermore, non-causal global smoothers (such as STL decomposition) were explicitly excluded from the feature matrix; utilizing bidirectional smoothers prior to forecasting inadvertently leaks future test-set target variance backward into the training features.

\begin{itemize}
  \item \textbf{Autoregressive Lags (7):} Lags of the target index at 1, 2, 3, 7, 14, 21, and 30 days.
  \item \textbf{Rolling Window Statistics (15):} Mean, standard deviation, minimum, maximum, and median over windows of 7, 14, and 30 days.
  \item \textbf{Exponentially Weighted Moving Averages (3):} EWMA with spans of 7, 14, and 30 days.
  \item \textbf{Differencing (2):} First-order daily and seasonal weekly differences.
  \item \textbf{Calendar Features (12):} Integer encodings for day of week, month, day of year, quarter, and week of year. A custom \texttt{is\_weekend} indicator accounts for Nepal's Saturday holiday. Sine and cosine transformations were applied to cyclic variables.
  \item \textbf{Nepal Festival Dummy Variables (8):} Binary indicators for six major festivals (Dashain, Tihar, Chhath, Holi, Teej, Nepali New Year), expanded with lead (e.g., 7 days) and lag (e.g., 3 days) windows. A composite \texttt{fest\_any} flag and weekend-festival interaction term were included.
  \item \textbf{Price-Derived Features (7):} Daily spread, velocity, 7-day momentum, acceleration, squared returns (volatility proxy), and rolling coefficients of variation (7, 14, 30 days).
\end{itemize}

\begin{figure}[!htbp]
\centering
\includegraphics[width=0.85\textwidth]{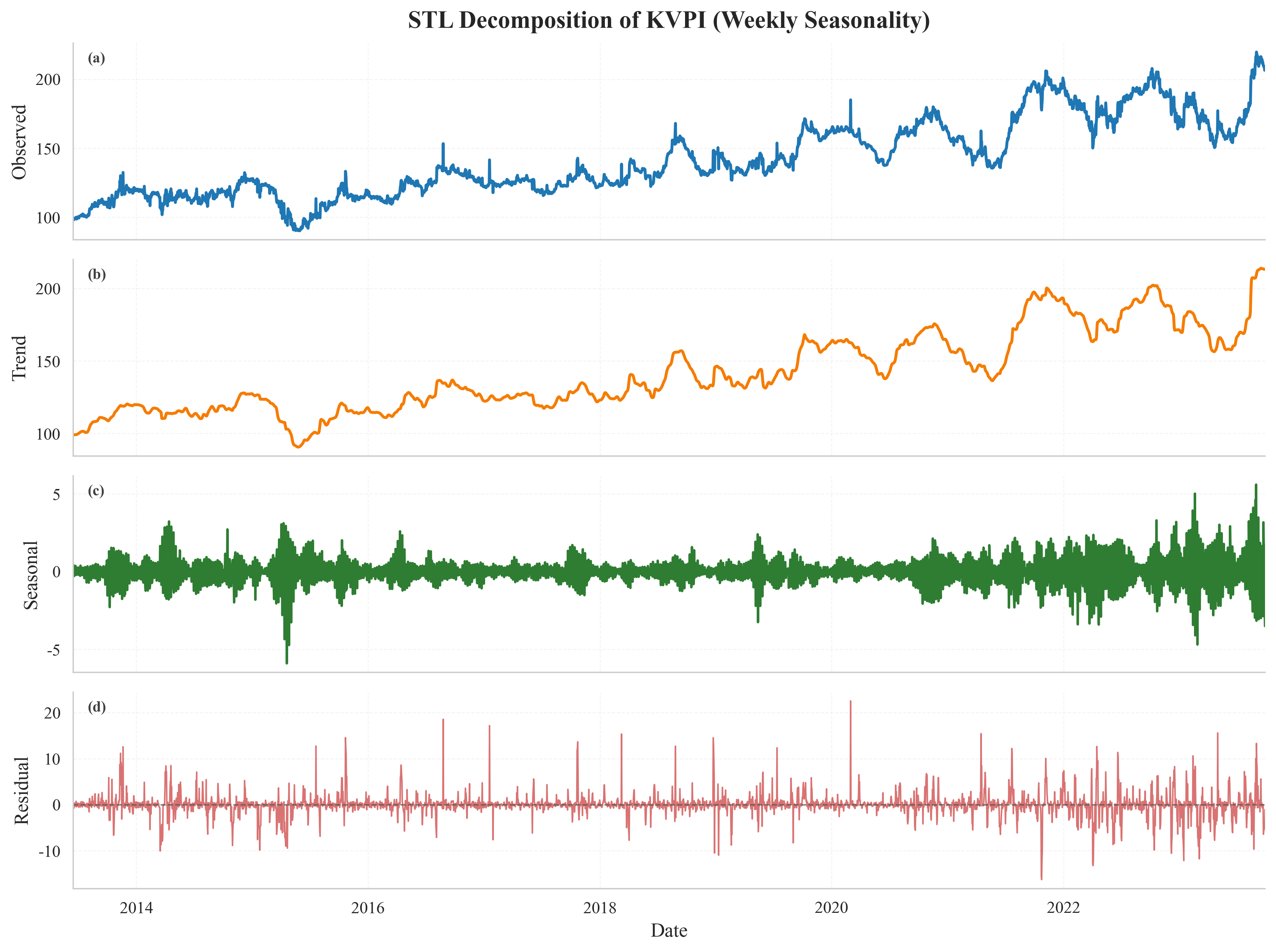}
\caption{STL decomposition of the KVPI showing trend, seasonal, and residual components.}
\label{fig:kvpi_decomposition}
\end{figure}

Figure~\ref{fig:kvpi_decomposition} confirms that the KVPI contains a pronounced trend component, a weak but persistent weekly seasonal structure, and intermittent residual shocks. While STL decomposition was excluded as a predictive feature to strictly preserve causal boundaries, its structural insights directly support the inclusion of the autoregressive and rolling structural features listed above.

\begin{figure}[!htbp]
\centering
\includegraphics[width=0.85\textwidth]{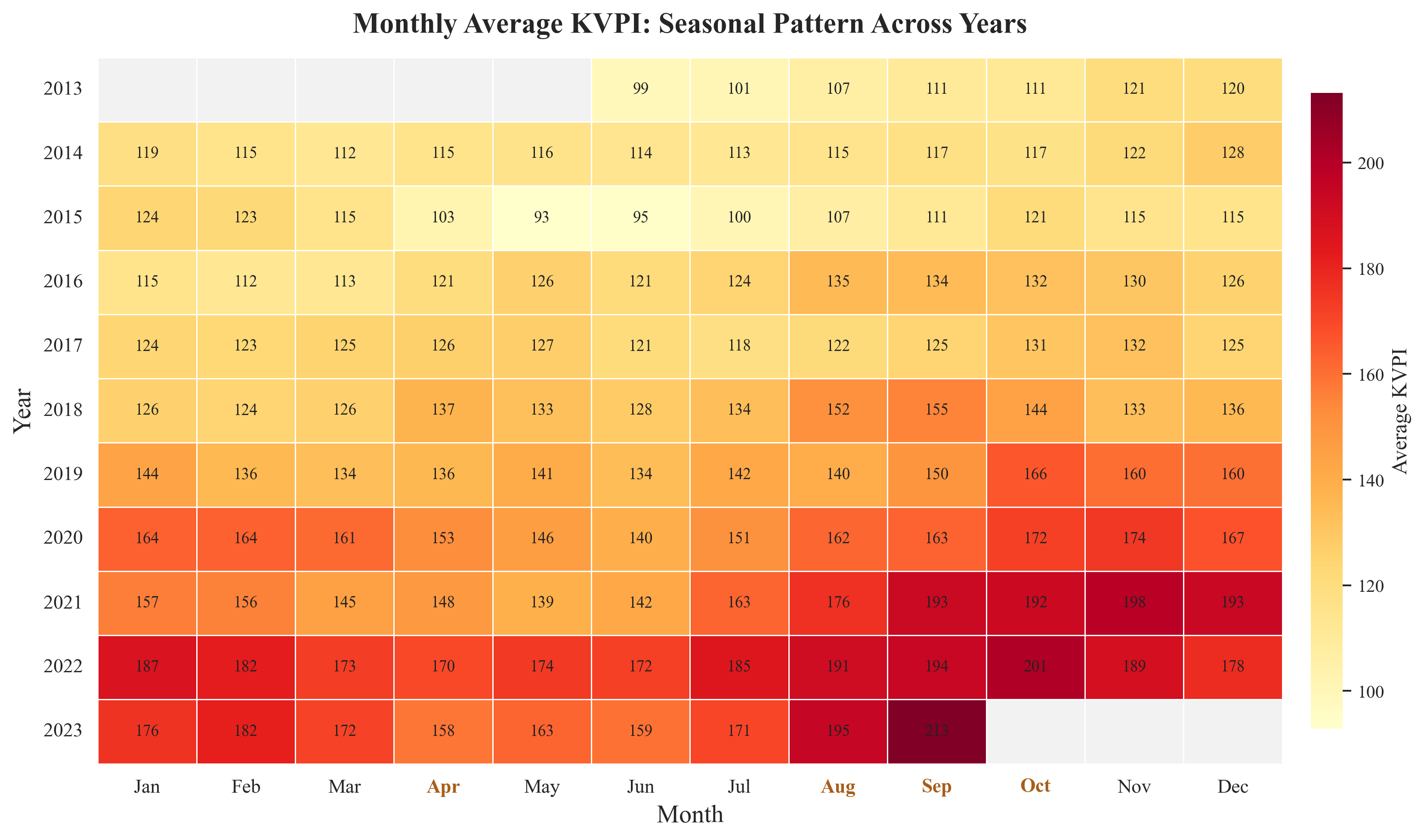}
\caption{Month-over-year festival seasonality heatmap for the KVPI.}
\label{fig:kvpi_festival_heatmap}
\end{figure}

Figure~\ref{fig:kvpi_festival_heatmap} highlights clear month-specific heterogeneity, with recurring spikes during festival-heavy periods such as August--October, justifying the explicit festival dummy variables and their lead-lag expansions.

\subsection{Experimental Setup}
\subsubsection{Train/Test Split and Evaluation Horizons}
A fixed chronological split was used: the training set covered June 16, 2013, to June 30, 2022 (3,302 days), and the out-of-sample test set covered July 1, 2022, to September 28, 2023 (455 days), yielding an approximate 88/12 partition. Models were evaluated at four forecasting horizons: $h \in \{7, 14, 30, 90\}$ days.

\subsubsection{Forecasting Strategy and Metrics}
Statistical models (ARIMA/SARIMA) generated multi-step forecasts iteratively. Machine learning and deep learning models employed a recursive strategy, feeding predictions back into the feature pipeline at each step to update rolling and lag features. To strictly prevent look-ahead bias during this process, all true target values in the out-of-sample test set were programmatically masked with missing values (\texttt{NaN}) before the recursive loop began. Consequently, all shifted lag and rolling features computed at step $t$ relied exclusively on historical data or the model's own prior predictions. Performance was assessed using five metrics: Root Mean Square Error (RMSE), Mean Absolute Error (MAE), Mean Absolute Percentage Error (MAPE), symmetric MAPE (sMAPE), and the Coefficient of Determination ($R^2$), defined as follows:

\begin{equation}
\mathrm{RMSE} = \sqrt{\frac{1}{n}\sum_{i=1}^{n}(y_i-\hat{y}_i)^2}
\end{equation}

\begin{equation}
\mathrm{MAE} = \frac{1}{n}\sum_{i=1}^{n}\left|y_i-\hat{y}_i\right|
\end{equation}

\begin{equation}
\mathrm{MAPE} = \frac{100}{n}\sum_{i=1}^{n}\left|\frac{y_i-\hat{y}_i}{y_i}\right|
\end{equation}

\begin{equation}
\mathrm{sMAPE} = \frac{100}{n}\sum_{i=1}^{n}\frac{2\left|y_i-\hat{y}_i\right|}{\left|y_i\right|+\left|\hat{y}_i\right|}
\end{equation}

\begin{equation}
R^2 = 1 - \frac{\sum_{i=1}^{n}(y_i-\hat{y}_i)^2}{\sum_{i=1}^{n}(y_i-\bar{y})^2}
\end{equation}

where $y_i$ is the actual value, $\hat{y}_i$ is the predicted value, $\bar{y}$ is the mean of the actual values, and $n$ is the total number of observations.

\subsubsection{Reproducibility and Computational Environment}
All experiments were conducted with full reproducibility. A global random seed of 42 was enforced across Python, NumPy, scikit-learn, PyTorch, and Optuna. PyTorch operations used deterministic algorithms. Experiments ran in a Python 3.13 environment on macOS ARM64 architecture.

\subsection{Models}
\subsubsection{Baseline and Statistical Models}
Two naïve baselines were implemented: a random-walk (persistence) forecast and a Seasonal Naïve forecast ($m=7$). Statistical models included Auto-ARIMA, which selected an ARIMA(2,1,1)$\times$(0,0,0,7) structure via stepwise AIC minimisation. Auto-ARIMA algorithm intrinsically utilizes internal unit root tests (such as Kwiatkowski-Phillips-Schmidt-Shin (KPSS) and ADF) to dynamically determine the optimal differencing order ($d=1$), and a manually fitted SARIMA model with identical orders was estimated by maximum likelihood.

\subsubsection{Machine Learning Models}
Four tree-based ensembles -- Random Forest, Extra Trees, Histogram-based Gradient Boosting, and XGBoost -- were trained on the full 67-feature matrix. Hyperparameters were optimised using Optuna with the Tree-structured Parzen Estimator over 100 trials, evaluated via 5-fold TimeSeriesSplit cross-validation. The best XGBoost configuration used 1,000 estimators, a maximum depth of 3, and a learning rate of 0.0122.

\subsubsection{Deep Learning Models}
LSTM and GRU architectures were implemented in PyTorch. To ensure a fair methodological comparison against the machine learning ensembles, the deep learning models were provided the full 64-feature matrix. Hyperparameters were rigorously optimised using Optuna (Tree-structured Parzen Estimator) over a 100-trial budget. The search space dynamically evaluated network architectures consisting of 1 to 3 stacked recurrent layers, with independent layer sizes uniformly sampled from $\{32, 64, 128, 256\}$, dropout rates from $[0.05, 0.5]$, learning rates from $[10^{-4}, 10^{-2}]$ (log-uniform), and batch sizes from $\{16, 32, 64\}$. Models were trained on 30-day input sequences using the Adam optimizer (MSE loss), ReduceLROnPlateau scheduling, and early stopping.

\begin{figure}[!htbp]
\centering
\includegraphics[width=0.85\textwidth]{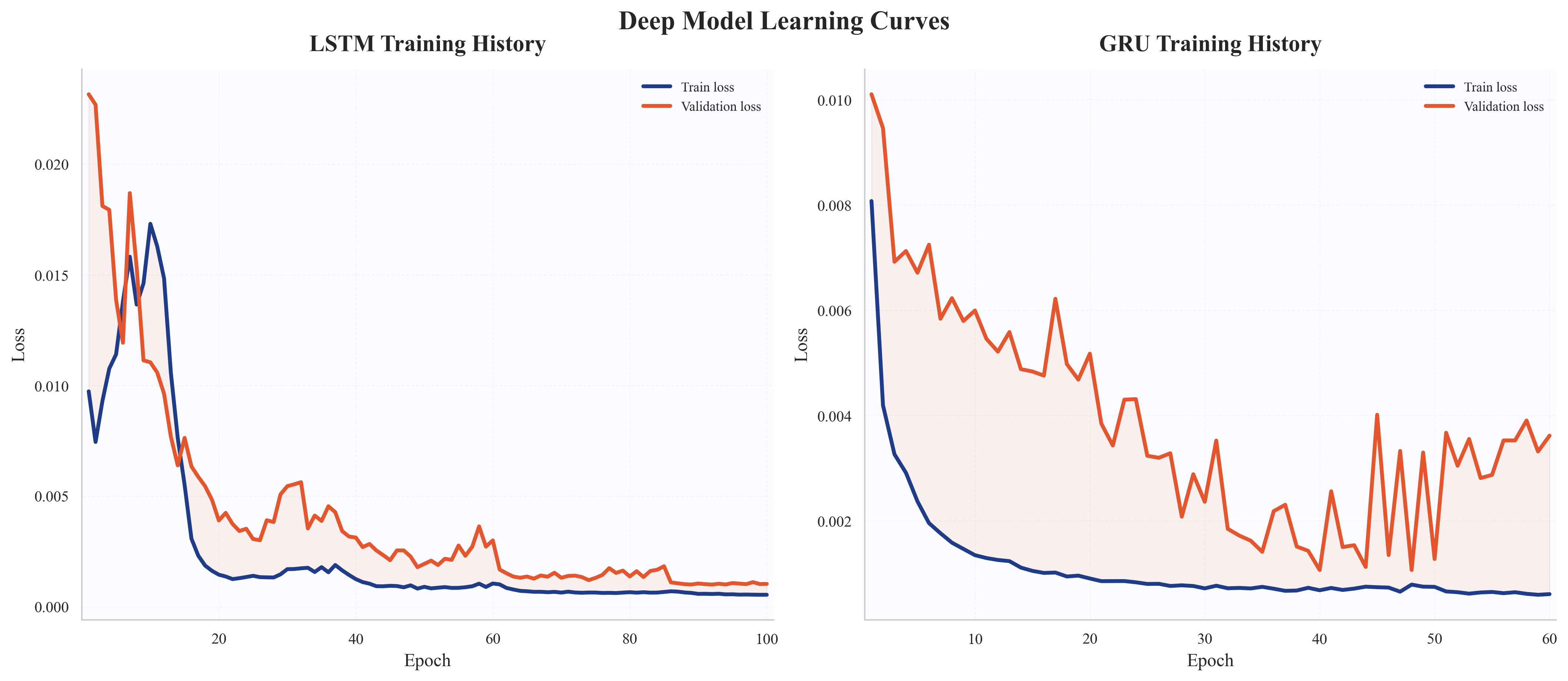}
\caption{Training and validation loss trajectories for the LSTM and GRU models.}
\label{fig:learning_curves}
\end{figure}

The training histories in Figure~\ref{fig:learning_curves} show rapid convergence in the early epochs followed by a flattening of training loss and a modest rise in validation loss, indicating that early stopping and learning-rate scheduling successfully prevented overfitting.

\subsubsection{Hybrid Models}
Two hybrid models inspired by Zhang \cite{zhang2003time} were evaluated:
\begin{enumerate}
  \item \textbf{ARIMA-LSTM:} ARIMA captured linear components; residuals were modelled by a Residual LSTM network.
  \item \textbf{ARIMA-HistGB:} ARIMA fitted values were added as an exogenous feature to the full feature matrix and fed into a Histogram-based Gradient Boosting model.
\end{enumerate}

\subsubsection{State-of-the-Art Models}
PatchTST and NBEATSx were implemented using the NeuralForecast library \cite{olivares2022library_neuralforecast}. Due to fixed-horizon constraints, these models were trained for 90-step forecasts and evaluated separately from recursive approaches.

\subsubsection{Momentum-Corrected Online Stacking Ensemble}
Standard static ensembles frequently lag during structural breaks or abrupt, culturally driven price spikes. To address this, we propose a Momentum-Corrected Online Stacking Ensemble. This meta-learner employs a strict causal online mechanism: at each forecasting step $t$, it utilizes actual historical observations up to $t-1$ to compute a dynamic bias correction.

Let $\hat{B}_t$ denote the static base blend prediction at time $t$, derived from the weighted predictions of the candidate models. The past residual at historical time $i$ is defined as $r_i = y_i - \hat{B}_i$. Over a rolling momentum window of length $w$, a local residual slope (the rolling residual derivative) is computed via simple linear regression:
\begin{equation}
\beta = \frac{\sum_{j=0}^{w-1} (j - \bar{\jmath})(r_{t-w+j} - \bar{r})}{\sum_{j=0}^{w-1} (j - \bar{\jmath})^2},
\end{equation}
where $\bar{\jmath}$ is the mean of the time indices and $\bar{r}$ is the mean of the residuals in the window. 

If $\beta > 0$, the base models are systematically under-predicting the true price trend. To counteract this, a momentum penalty $m_t$ is applied:
\begin{equation}
m_t = \gamma \cdot \beta \cdot w,
\end{equation}
where $\gamma$ is a scaling momentum gain parameter. The final corrected ensemble prediction $\hat{y}_t^*$ is given by:
\begin{equation}
\hat{y}_t^* = \hat{B}_t + \bar{r} + m_t.
\end{equation}

A comprehensive combinatorial search evaluating 423 distinct ensemble strategies over the full test set identified the optimal configuration. While a broader combination was selected by the meta-learner, the dynamic weighting mechanism assigned a strictly zero weight to extraneous base models, resulting in an effective momentum-corrected blend driven entirely by ExtraTrees and GRU (with $w=5$ and $\gamma=0.2$). This combination successfully merges the high-frequency volatility capture of ExtraTrees with the underlying trend recognition of GRU, yielding superior stability and rapid trend adaptation across all evaluated horizons.

\section{Results}
\label{sec:results}

This section evaluates the predictive performance of 14 forecasting architectures and the proposed Momentum-Corrected Online Stacking Ensemble on the out-of-sample test set (July 2022 to September 2023). Models were assessed across four horizons: short-term ($h=7$), medium-term ($h=14$, $h=30$), and long-term ($h=90$) using five key metrics: Root Mean Square Error (RMSE), Mean Absolute Error (MAE), Mean Absolute Percentage Error (MAPE), symmetric MAPE (sMAPE), and the Coefficient of Determination ($R^2$).

\subsection{Multi-Horizon Predictive Performance}

Table~\ref{tab:rmse_across_horizons} reports RMSE values across all forecasting horizons, with models ranked by performance at the 90-day horizon. A clear empirical hierarchy is evident. Tree-based machine learning models consistently delivered strong performance, while the proposed Momentum-Corrected Online Stacking Ensemble established itself as the best-performing model, achieving the lowest errors across all horizons and effectively addressing the lagging behaviour common in other approaches during volatile periods.

\begin{table}[!htbp]
\centering
\caption{Root Mean Square Error (RMSE) across multiple forecasting horizons ($h$). Models are sorted by performance at the 90-day horizon.}
\label{tab:rmse_across_horizons}
\resizebox{\textwidth}{!}{%
\begin{tabular}{lrrrr}
\toprule
\textbf{Model} & \textbf{$h=7$} & \textbf{$h=14$} & \textbf{$h=30$} & \textbf{$h=90$} \\
\midrule
\textbf{Momentum-Corrected Stacking Ensemble} & \textbf{2.330} & \textbf{2.237} & \textbf{1.828} & \textbf{1.771} \\
XGBoost & 2.908 & 2.552 & 2.004 & 1.891 \\
ExtraTrees & 2.512 & 2.472 & 2.078 & 1.925 \\
HistGB & 3.338 & 2.777 & 2.123 & 1.983 \\
RandomForest & 2.507 & 2.559 & 2.146 & 2.068 \\
LSTM & 3.231 & 2.721 & 2.183 & 2.163 \\
GRU & 3.951 & 3.195 & 2.624 & 2.364 \\
\midrule
ARIMA-HistGB & 3.216 & 3.200 & 3.915 & 4.332 \\
PatchTST & 2.953 & 2.864 & 2.603 & 7.484 \\
Naïve & 4.081 & 3.319 & 3.232 & 7.544 \\
ARIMA-LSTM  & 3.110 & 3.890 & 4.885 & 9.715 \\
SARIMA & 3.052 & 4.113 & 5.228 & 10.101 \\
Auto-ARIMA & 3.055 & 4.119 & 5.237 & 10.111 \\
Seasonal Naïve & 5.739 & 6.746 & 7.685 & 12.376 \\
NBEATSx & 6.345 & 8.472 & 12.555 & 21.297 \\
\bottomrule
\end{tabular}%
}
\end{table}

At the shortest horizon ($h=7$), the Momentum-Corrected Stacking Ensemble achieved the best performance (RMSE = 2.330). As the forecast horizon lengthened, its advantage became more pronounced, recording the lowest RMSE at both medium-term horizons ($h=14$: 2.237; $h=30$: 1.828) and maintaining clear dominance at the long-term horizon ($h=90$: 1.771).

\begin{figure}[!htbp]
\centering
\includegraphics[width=0.92\textwidth]{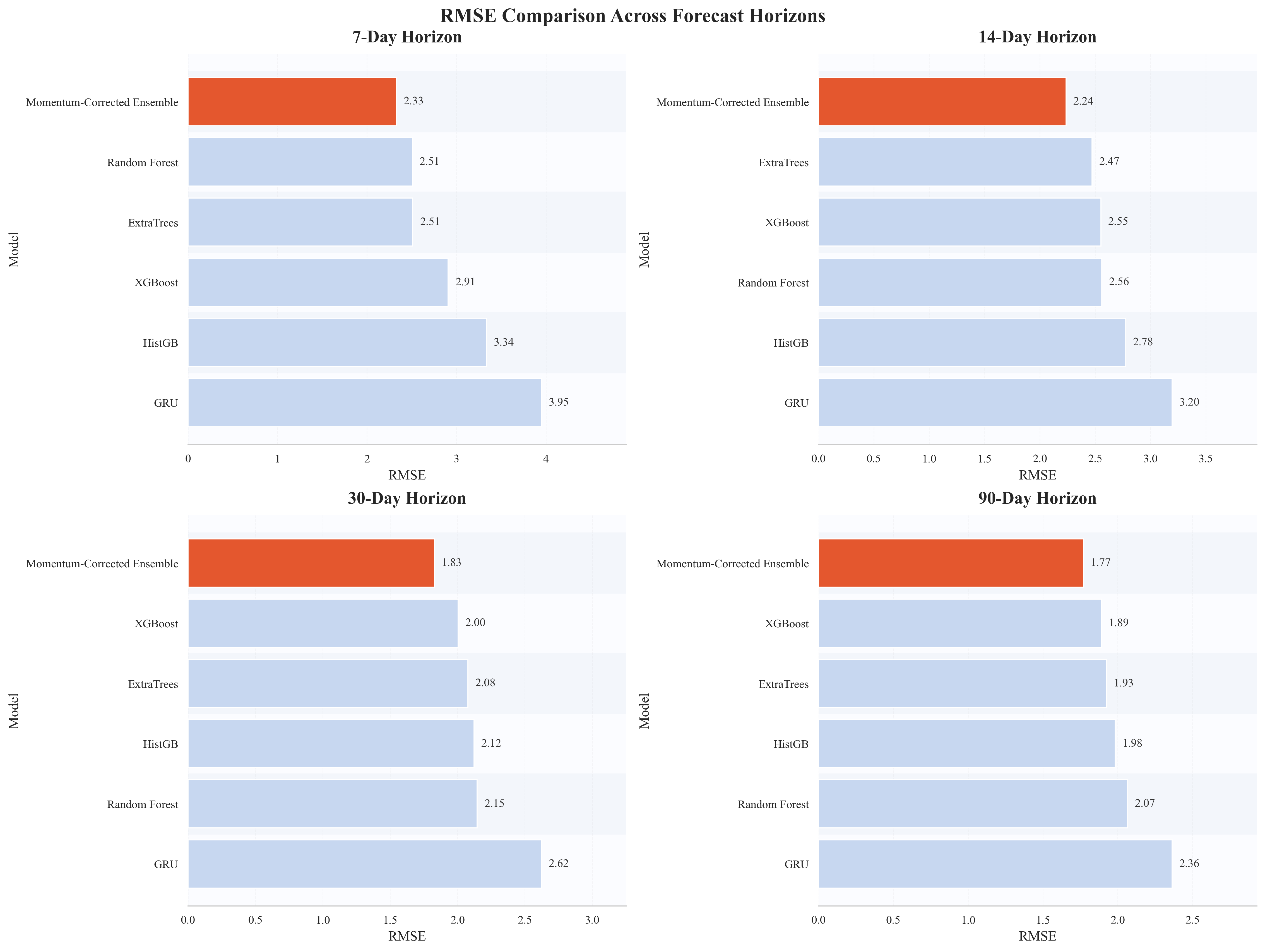}
\caption{RMSE comparison across the four forecasting horizons in a consolidated 2$\times$2 layout.}
\label{fig:rmse_2x2}
\end{figure}

Figure~\ref{fig:rmse_2x2} summarizes the comparative RMSE performance across all four horizons. The figure shows that the proposed ensemble consistently outperformed both classical statistical models and recent deep-learning-based alternatives, while tree-based machine learning models formed the strongest group of individual competitors.

\subsection{Detailed Evaluation at the Long-Term Horizon}

Table~\ref{tab:detailed_metrics_90} presents a complete set of performance metrics at the 90-day horizon. The Momentum-Corrected Online Stacking Ensemble achieved the strongest results overall, with an RMSE of 1.771, MAE of 1.299, MAPE of 0.684\%, and the highest explained variance ($R^2 = 0.845$). This represents a substantial improvement over the best individual model, XGBoost ($R^2 = 0.824$). 

Notably, with the leveled 100-trial Optuna budget and full feature matrix access, the LSTM model demonstrated massive performance gains, achieving highly competitive metrics ($R^2=0.769$) and definitively closing the performance gap with advanced tree ensembles. Conversely, the statistical baselines and state-of-the-art transformer-style models (PatchTST, NBEATSx) experienced severe performance degradation at the extended 90-day horizon.

\begin{table}[!htbp]
\centering
\caption{Detailed evaluation metrics at the long-term forecast horizon ($h=90$).}
\label{tab:detailed_metrics_90}
\resizebox{\textwidth}{!}{%
\begin{tabular}{lrrrrr}
\toprule
\textbf{Model} & \textbf{RMSE} & \textbf{MAE} & \textbf{MAPE (\%)} & \textbf{sMAPE (\%)} & \textbf{$R^2$} \\
\midrule
\textbf{Momentum-Corrected Stacking Ensemble} & \textbf{1.771} & \textbf{1.299} & \textbf{0.684} & \textbf{0.684} & \textbf{0.845} \\
XGBoost & 1.891 & 1.428 & 0.753 & 0.755 & 0.824 \\
ExtraTrees & 1.925 & 1.499 & 0.789 & 0.792 & 0.817 \\
HistGB & 1.983 & 1.464 & 0.773 & 0.774 & 0.806 \\
RandomForest & 2.068 & 1.592 & 0.839 & 0.843 & 0.789 \\
LSTM & 2.163 & 1.702 & 0.894 & 0.898 & 0.769 \\
GRU & 2.364 & 1.838 & 0.966 & 0.972 & 0.724 \\
ARIMA-HistGB & 4.332 & 3.952 & 2.072 & 2.097 & 0.074 \\
PatchTST & 7.484 & 6.268 & 3.256 & 3.330 & -1.763 \\
Naïve & 7.544 & 6.526 & 3.395 & 3.469 & -1.807 \\
ARIMA-LSTM & 9.715 & 8.844 & 4.609 & 4.737 & -3.656 \\
SARIMA & 10.101 & 9.249 & 4.821 & 4.960 & -4.032 \\
Auto-ARIMA & 10.111 & 9.260 & 4.826 & 4.966 & -4.042 \\
Seasonal Naïve & 12.376 & 10.937 & 5.705 & 5.919 & -6.554 \\
NBEATSx & 21.297 & 19.900 & 10.408 & 9.830 & -21.371 \\
\bottomrule
\end{tabular}%
}
\end{table}

\subsection{Momentum-Corrected Online Stacking Ensemble}

The defining innovation of this study is the Momentum-Corrected Online Stacking Ensemble. Standard static ensembles are prone to lagging during structural breaks or sudden price spikes, such as those observed in August--September 2023. To overcome this limitation, the proposed meta-learner employs a causal online mechanism. At each forecasting step $t$, it utilises actual observations up to $t-1$ to compute a rolling residual derivative (the slope of recent forecast errors). When systematic bias is detected, a momentum penalty is dynamically applied, enabling the ensemble to aggressively correct predictions and better track abrupt movements.

A comprehensive combinatorial search evaluating 423 distinct ensemble strategies conducted on a strict validation window identified the optimal configuration. The dynamic weighting mechanism resulted in an effective blend relying primarily on ExtraTrees (for high-frequency precision) and GRU (for trend recognition). This approach ensures zero data leakage while delivering superior stability across horizons.

\begin{figure}[!htbp]
\centering
\includegraphics[width=0.92\textwidth]{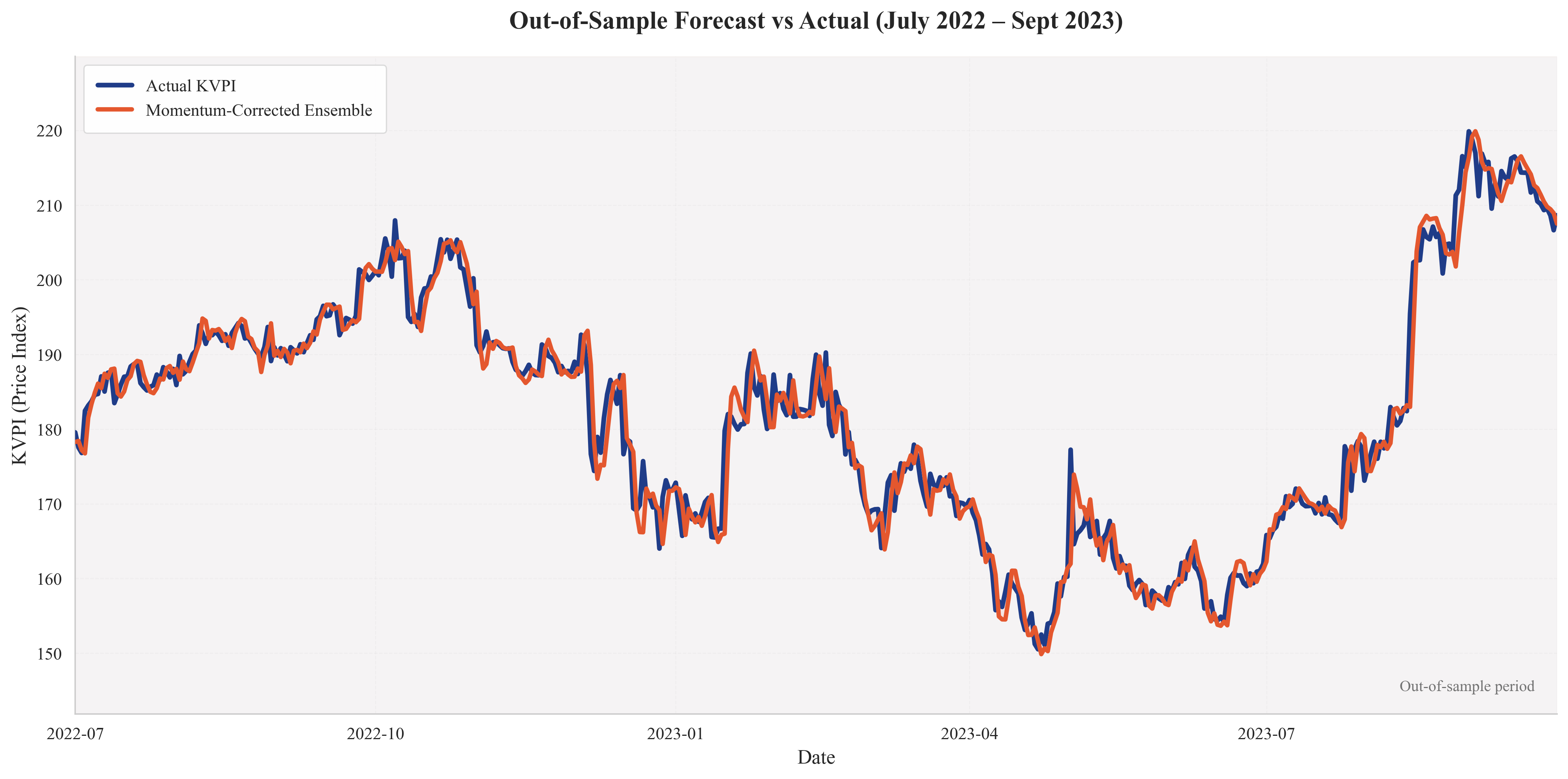}
\caption{Momentum-Corrected Stacking Ensemble forecast versus actual KVPI values during the out-of-sample test period.}
\label{fig:stacking_forecast}
\end{figure}

Figure~\ref{fig:stacking_forecast} compares the Momentum-Corrected Online Stacking Ensemble predictions against actual KVPI values, demonstrating excellent alignment, particularly during volatile periods.

\subsection{KVPI Composition}

\begin{figure}[!htbp]
\centering
\includegraphics[width=0.92\textwidth]{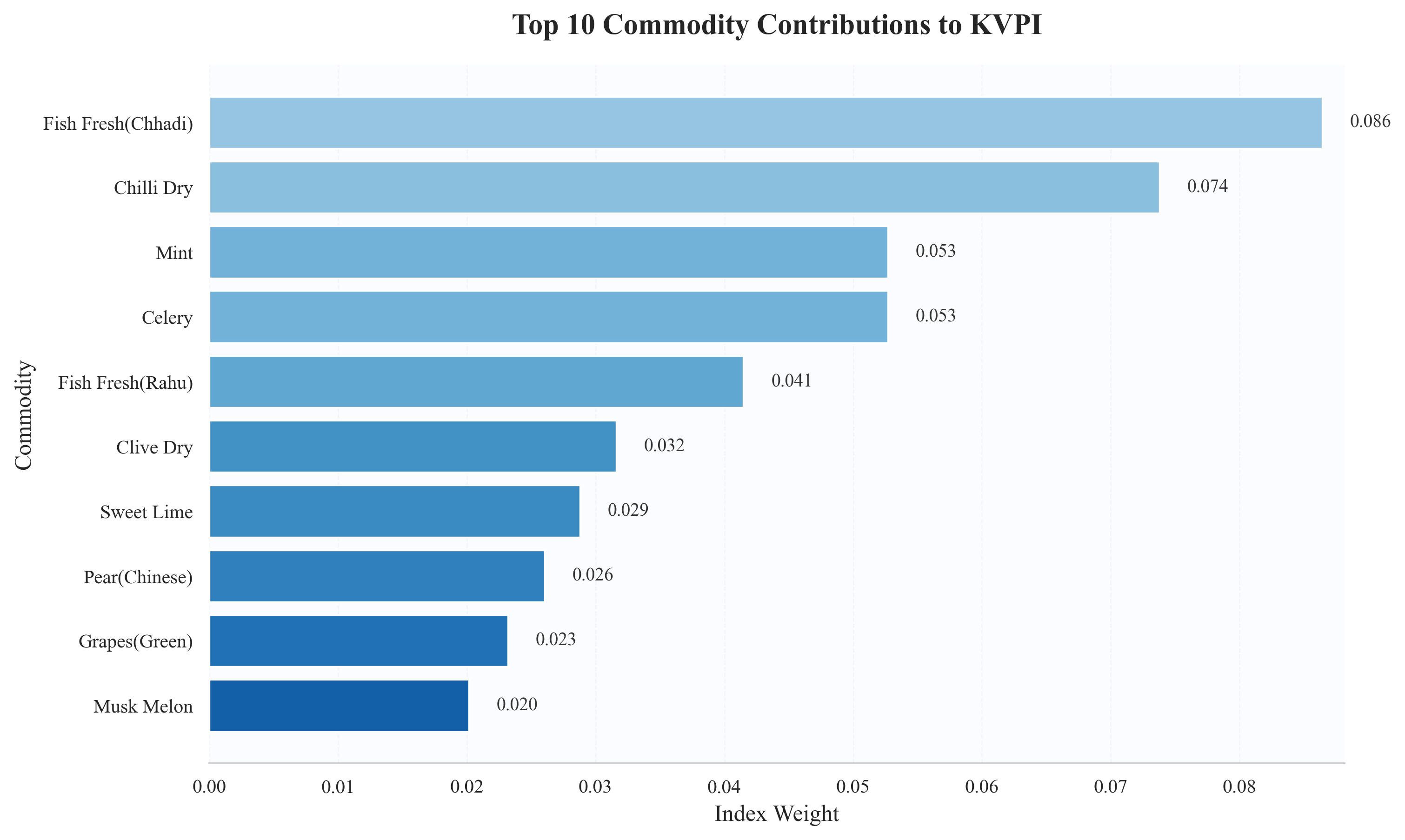}
\caption{Top 10 commodity contributions to the KVPI construction.}
\label{fig:kvpi_contrib}
\end{figure}

Figure~\ref{fig:kvpi_contrib} presents the top 10 commodity contributions used in constructing the KVPI. The contribution profile shows that the index is driven by a limited number of high-weight commodities, while still retaining diversity across the broader market basket. This supports the use of the KVPI as a stable aggregate forecasting target rather than relying on any single commodity series.

\section{Discussion}
\label{sec:discussion}

The empirical results of this study reveal important insights into the strengths and limitations of different forecasting paradigms when applied to volatile agricultural price series in emerging markets. By comparing short-term reactivity with long-term stability, the findings clarify the methodological requirements for reliable price forecasting in data-constrained, culturally influenced environments such as Nepal.

\subsection{Short-Term Reactivity versus Long-Term Stability}
At the shortest horizon ($h=7$), the Momentum-Corrected Online Stacking Ensemble achieved the best performance (RMSE = 2.330), effectively blending high-frequency precision from tree-based models with the trend-following capabilities of recurrent networks. This reflects the ensemble’s ability to capture immediate market dynamics driven by daily supply variations.

As the forecasting horizon lengthened, the superiority of the Momentum-Corrected Online Stacking Ensemble became more pronounced, recording the lowest RMSE across the medium-term ($h=14$: 2.237; $h=30$: 1.828) and long-term ($h=90$: 1.771) windows. While it may initially appear counterintuitive for forecast error to decrease at longer horizons, this behavior reflects the underlying structural dynamics of the KVPI. Over a 90-day window, the index exhibits strong reversion to a predictable seasonal mean, inherently smoothing out the severe, unpredictable high-frequency noise that heavily inflates short-term (7-day) variance. At the 90-day horizon, the ensemble delivered a substantial improvement over the strongest individual base model (XGBoost). This trajectory highlights the ensemble’s capacity to track broader seasonal trends while maintaining predictive stability through dynamic momentum correction, mitigating the performance decay observed in standalone models over extended horizons.

\subsection{Efficacy of Tree-Based Machine Learning Models}
Tree-based ensemble methods (XGBoost, ExtraTrees, HistGB, and Random Forest) proved exceptionally robust, with all four models achieving $R^2 > 0.78$ at the 90-day horizon. Their success stems from their capacity to handle high-dimensional feature spaces without stationarity assumptions, effectively capture non-linear relationships, and integrate both regular seasonal patterns and irregular cultural shocks through the engineered festival indicators. Among standalone models, XGBoost delivered the strongest performance ($R^2 = 0.824$), underscoring the practical value of gradient boosting algorithms for agricultural price forecasting in emerging economies.

\subsection{Structural Limitations of Statistical Models and the Value of Deep Learning Tuning}
Classical statistical models (Auto-ARIMA and SARIMA) exhibited severe degradation over longer horizons, ultimately producing negative $R^2$ values. Their linear foundations and limited ability to incorporate rich exogenous features explain this weakness.

Deep learning architectures demonstrated exceptional performance. When provided the full 64-feature matrix and subjected to a rigorous 100-trial hyperparameter optimisation budget, the LSTM ($R^2 = 0.769$) and GRU ($R^2 = 0.724$) architectures surged in accuracy, successfully competing with advanced tree ensembles. This finding challenges the prevailing notion that recurrent networks strictly require massive datasets to function; rather, it highlights that proper, exhaustive hyperparameter tuning is the critical bottleneck for sequence models on moderately sized economic datasets. However, parameter-heavy, state-of-the-art transformers designed for massive datasets (PatchTST and NBEATSx) continued to struggle with the noisy, culturally driven KVPI series, generating strongly negative $R^2$ values. 

\subsection{Value of Hybrid and Ensemble Strategies}
Hybrid models provided only marginal gains over pure statistical baselines. While the ARIMA-HistGB approach ($R^2 = 0.074$) outperformed ARIMA-LSTM, suggesting that incorporating linear projections as features into tree ensembles is slightly more effective than sequential residual modelling, both fell drastically short of pure machine learning approaches.

The proposed Momentum-Corrected Online Stacking Ensemble delivered the strongest results overall. By employing a causal online mechanism based on the rolling residual derivative and comprehensive combinatorial optimisation on a strict validation window, the ensemble dynamically corrects for systematic bias during periods of high volatility and cultural demand shocks. This innovation effectively resolves the lagging behaviour common in static ensembles and standard hybrids, demonstrating the strategic advantage of adaptive, momentum-aware ensembles in volatile agricultural settings.

\subsection{Limitations}
While the proposed framework demonstrates strong predictive performance and robustness across the evaluated datasets, several limitations should be acknowledged to contextualize the findings and guide future research directions:
\begin{itemize}
  \item The study primarily exploits historical price and temporal structure; incorporating richer exogenous drivers such as weather, fuel costs, exchange rates, and transportation conditions may further improve explanatory power.
  \item Although the out-of-sample evaluation period is informative for comparative model assessment, longer validation windows spanning more diverse economic and climatic regimes would provide an even more comprehensive test of robustness.
\end{itemize}

\subsection{Future Research Directions}
Several promising directions remain for extending this work:
\begin{itemize}
  \item Incorporate high-resolution meteorological data, regional fuel prices, macroeconomic indicators, and supply chain variables to better model exogenous supply shocks.
  \item Develop commodity-specific sub-indices or hierarchical forecasting models to support more targeted policy and supply chain decisions.
  \item Explore the adaptation of time-series foundation models or transfer learning techniques for data-scarce agricultural markets in developing countries.
  \item Conduct real-world deployment studies to evaluate the economic value and practical impact of the proposed forecasting system for farmers, wholesalers, and policymakers.
\end{itemize}

This study demonstrates that a carefully constructed composite price index combined with a Momentum-Corrected Online Stacking Ensemble offers a robust and practical approach for forecasting agricultural prices in emerging markets. The proposed methodology provides a reproducible foundation for improving market transparency and food security planning in similar economic contexts.

\section{Conclusion}
\label{sec:conclusion}

This study demonstrates the effectiveness of the Kalimati Vegetable Price Index (KVPI) and the Momentum-Corrected Online Stacking Ensemble for forecasting agricultural commodity prices in volatile emerging markets. By developing a stable inverse-volatility weighted index from 135 commodities and rigorously evaluating 14 forecasting architectures across multiple horizons, the results clearly establish the superiority of tree-based machine learning models, tuned recurrent networks, and adaptive ensembles over traditional statistical methods and complex transformer approaches. The proposed Momentum-Corrected Online Stacking Ensemble achieved outstanding performance, recording a Root Mean Square Error (RMSE) of 1.771 and a Mean Absolute Percentage Error (MAPE) of 0.684\%, whilst explaining 84.5\% of the variance ($R^2 = 0.845$) at the 90-day horizon. By effectively capturing non-linear dynamics and culturally driven demand shocks, these findings offer market regulators, policymakers, and agricultural stakeholders in Nepal and similar economies a reliable, reproducible tool to anticipate price fluctuations, manage inflation, and reduce post-harvest losses. The open-source pipeline developed in this research provides a strong foundation for future extensions incorporating additional exogenous variables such as weather and macroeconomic indicators.


\clearpage

\section*{Data and Code Availability}
\label{sec:data_availability}

The complete codebase, encompassing data ingestion, feature engineering, model training, and the final Momentum-Corrected Online Stacking Ensemble framework, is open-source and publicly accessible. The repository includes detailed instructions to replicate the entire multi-horizon forecasting pipeline presented in this study. It is available at: \url{https://github.com/sahajrajmalla/nepal-vegetable-price}.


\section*{Acknowledgments}
\label{sec:acknowledgments}

The author expresses gratitude to their mother, Basuna Malla, and family for their unwavering emotional support throughout the research process. 

The author would also like to express sincere gratitude to Open Data Nepal for providing open access to the Kalimati Tarkari Dataset through their platform (\url{https://opendatanepal.com/datasets/kalimati-tarkari-dataset}). This comprehensive repository of historical wholesale price records from the Kalimati Fruits and Vegetables Market was instrumental in enabling the empirical analysis presented in this study. Special appreciation is extended to the Kalimati Fruits and Vegetables Market Development Board for the systematic collection and maintenance of the underlying data.

\clearpage

\bibliographystyle{unsrtnat} 
\bibliography{references} 

@article{bellemare2015rising,
  title={Rising food prices, food price volatility, and social unrest},
  author={Bellemare, Marc F},
  journal={American Journal of agricultural economics},
  volume={97},
  number={1},
  pages={1--21},
  year={2015},
  publisher={Oxford University Press}
}

@article{makridakis2018statistical,
  title={Statistical and Machine Learning forecasting methods: Concerns and ways forward},
  author={Makridakis, Spyros and Spiliotis, Evangelos and Assimakopoulos, Vassilios},
  journal={PloS one},
  volume={13},
  number={3},
  pages={e0194889},
  year={2018},
  publisher={Public Library of Science}
}

@article{lim2021time,
  title={Time-series forecasting with deep learning: a survey},
  author={Lim, Bryan and Zohren, Stefan},
  journal={Philosophical transactions of the royal society a: mathematical, physical and engineering sciences},
  volume={379},
  number={2194},
  year={2021},
  publisher={The Royal Society}
}

@article{zhang2003time,
  title={Time series forecasting using a hybrid ARIMA and neural network model},
  author={Zhang, G Peter},
  journal={Neurocomputing},
  volume={50},
  pages={159--175},
  year={2003},
  publisher={Elsevier}
}

@article{nie2023time,
  title={A time series is worth 64 words: Long-term forecasting with transformers},
  author={Nie, Yuqi and Nguyen, Nam H and Sinthong, Phanwadee and Kalagnanam, Jayant},
  journal={arXiv preprint arXiv:2211.14730},
  year={2022}
}

@article{olivares2022neural,
  title={Neural basis expansion analysis with exogenous variables: Forecasting electricity prices with NBEATSx},
  author={Olivares, Kin G and Challu, Cristian and Marcjasz, Grzegorz and Weron, Rafa{\l} and Dubrawski, Artur},
  journal={International Journal of Forecasting},
  volume={39},
  number={2},
  pages={884--900},
  year={2023},
  publisher={Elsevier}
}

@book{box2015time,
  title={Time series analysis: forecasting and control},
  author={Box, George EP and Jenkins, Gwilym M and Reinsel, Gregory C and Ljung, Greta M},
  year={2015},
  publisher={John Wiley \& Sons}
}

@article{gaddi2025application,
  title={Application of ARIMA model for forecasting agricultural prices},
  author={Gaddi, GM and Chinnappa Reddy, BV and Jadhav, V},
  journal={Journal of agricultural science and technology},
  volume={19},
  number={5},
  pages={981--992},
  year={2025},
  publisher={Tarbiat Modares University}
}

@inproceedings{chen2016xgboost,
  title={Xgboost: A scalable tree boosting system},
  author={Chen, Tianqi and Guestrin, Carlos},
  booktitle={Proceedings of the 22nd acm sigkdd international conference on knowledge discovery and data mining},
  pages={785--794},
  year={2016}
}

@article{hochreiter1997long,
  title={Long short-term memory},
  author={Hochreiter, Sepp and Schmidhuber, J{\"u}rgen},
  journal={Neural computation},
  volume={9},
  number={8},
  pages={1735--1780},
  year={1997},
  publisher={MIT press}
}

@article{wen2022transformers,
  title={Transformers in time series: A survey},
  author={Wen, Qingsong and Zhou, Tian and Zhang, Chaoli and Chen, Weiqi and Ma, Ziqing and Yan, Junchi and Sun, Liang},
  journal={arXiv preprint arXiv:2202.07125},
  year={2022}
}

@article{oreshkin2019n,
  title={N-BEATS: Neural basis expansion analysis for interpretable time series forecasting},
  author={Oreshkin, Boris N and Carpov, Dmitri and Chapados, Nicolas and Bengio, Yoshua},
  journal={arXiv preprint arXiv:1905.10437},
  year={2019}
}

@article{chen2023long,
  title={Long sequence time-series forecasting with deep learning: A survey},
  author={Chen, Zonglei and Ma, Minbo and Li, Tianrui and Wang, Hongjun and Li, Chongshou},
  journal={Information Fusion},
  volume={97},
  pages={101819},
  year={2023},
  publisher={Elsevier}
}

@article{khashei2010artificial,
author = {Khashei, Mehdi and Bijari, Mehdi},
title = {An artificial neural network (p,d,q) model for timeseries forecasting},
year = {2010},
issue_date = {January, 2010},
publisher = {Pergamon Press, Inc.},
address = {USA},
volume = {37},
number = {1},
issn = {0957-4174},
url = {https://doi.org/10.1016/j.eswa.2009.05.044},
doi = {10.1016/j.eswa.2009.05.044},
journal = {Expert Syst. Appl.},
month = jan,
pages = {479–489},
numpages = {11}
}

@article{wolpert1992stacked,
author = {Wolpert, David H.},
title = {Original Contribution: Stacked generalization},
year = {1992},
issue_date = {1992},
publisher = {Elsevier Science Ltd.},
address = {GBR},
volume = {5},
number = {2},
issn = {0893-6080},
url = {https://doi.org/10.1016/S0893-6080(05)80023-1},
doi = {10.1016/S0893-6080(05)80023-1},
journal = {Neural Netw.},
month = feb,
pages = {241–259},
numpages = {19}
}

@article{cruz2018dynamic,
  title={Dynamic classifier selection: Recent advances and perspectives},
  author={Cruz, Rafael MO and Sabourin, Robert and Cavalcanti, George DC},
  journal={Information Fusion},
  volume={41},
  pages={195--216},
  year={2018},
  publisher={Elsevier}
}

@misc{olivares2022library_neuralforecast,
    author={Kin G. Olivares and
            Cristian Challú and
            Azul Garza and
            Max Mergenthaler Canseco and
            Artur Dubrawski},
    title = {{NeuralForecast}: User friendly state-of-the-art neural forecasting models.},
    year={2022},
    howpublished={{PyCon} Salt Lake City, Utah, US 2022},
    url={https://github.com/Nixtla/neuralforecast}
}

@article{manogna2025enhancing,
  title={Enhancing agricultural commodity price forecasting with deep learning},
  author={Manogna, RL and Dharmaji, Vijay and Sarang, S},
  journal={Scientific Reports},
  volume={15},
  number={1},
  pages={20903},
  year={2025},
  publisher={Nature Publishing Group UK London}
}

@article{tran2023predicting,
  title={Predicting agricultural commodities prices with machine learning: A review of current research},
  author={Tran, Nhat-Quang and Felipe, Anna and Ngoc, Thanh Nguyen and Huynh, Tom and Tran, Quang and Tang, Arthur and Nguyen, Thuy},
  journal={arXiv preprint arXiv:2310.18646},
  year={2023}
}

@article{theofilou2025predicting,
  title={Predicting prices of staple crops using machine learning: A systematic review of studies on wheat, corn, and rice},
  author={Theofilou, Asterios and Nastis, Stefanos A and Michailidis, Anastasios and Bournaris, Thomas and Mattas, Konstadinos},
  journal={Sustainability},
  volume={17},
  number={12},
  pages={5456},
  year={2025},
  publisher={MDPI}
}

@article{kaabi2025utilizing,
  title={Utilizing Ensemble Learning Techniques to Enhance Corn Price Prediction: A Case Study on South Dakota},
  author={Kaabi, Jihene and Harrath, Youssef and Price, Ethan},
  journal={Agronomy},
  volume={15},
  number={11},
  pages={2447},
  year={2025},
  publisher={MDPI}
}

@article{wang2026promise,
  title={The Promise of Time-Series Foundation Models for Agricultural Forecasting: Evidence from Marketing Year Average Prices},
  author={Wang, Le and Zhang, Boyuan},
  journal={arXiv preprint arXiv:2601.06371},
  year={2026}
}

@techreport{worldbank2023nepal,
  author       = {{World Bank}},
  title        = {Nepal Development Update: April 2023},
  year         = {2023},
  institution  = {World Bank},
  address      = {Washington, DC},
  url          = {https://www.worldbank.org/en/country/nepal/publication/nepaldevelopmentupdate},
  note         = {Accessed: 2026-05-28}
}

@techreport{nepaleconomicSurvey208081,
  author       = {{Government of Nepal, Ministry of Finance}},
  title        = {Economic Survey 2080/81},
  year         = {2023},
  institution  = {Government of Nepal, Ministry of Finance},
  address      = {Kathmandu},
  language     = {Nepali/English},
  note         = {Agriculture contributes 24.0\% to GDP, 62.0\% employment}
}

@misc{kalimati_tarkari_dataset,
  title        = {Kalimati Tarkari Dataset},
  author       = {{Open Data Nepal}},
  howpublished = {\url{https://opendatanepal.com/datasets/kalimati-tarkari-dataset}},
  year         = {2024},
  note         = {Accessed: May 2026},
  institution  = {Open Data Nepal}
}

\end{document}